# DETECTING DANGER: APPLYING A NOVEL IMMUNOLOGICAL CONCEPT TO INTRUSION DETECTION SYSTEMS

Julie Greensmith, Uwe Aickelin & Jamie Twycross
ASAP Group, School Of Computer Science,
University Of Nottingham, UK.
{jqg, uxa, jpt}@cs.nott.ac.uk

## INTRODUCTION

In recent years computer systems have become increasingly complex and consequently the challenge of protecting these systems has become increasingly difficult. Various techniques have been implemented to counteract the misuse of computer systems in the form of firewalls, anti-virus software and intrusion detection systems. The complexity of networks and dynamic nature of computer systems leaves current methods with significant room for improvement.

Computer scientists have recently drawn inspiration from mechanisms found in biological systems and, in the context of computer security, have focused on the human immune system (HIS). The human immune system provides an example of a robust, distributed system that provides a high level of protection from constant attacks. By examining the precise mechanisms of the human immune system, it is hoped the paradigm will improve the performance of real intrusion detection systems.

This paper presents an introduction to recent developments in the field of immunology. It discusses the incorporation of a novel immunological paradigm, Danger Theory, and how this concept is inspiring artificial immune systems (AIS). Applications within the context of computer security are outlined drawing direct reference to the underlying principles of Danger Theory and finally, the current state of intrusion detection systems is discussed and improvements suggested.

## DANGER THEORY AND THE HUMAN IMMUNE SYSTEM

Since 1959, the central dogma of immunology has stated that the human immune system reacts to entities that are not part of the organism. Therefore the decision to react is a result of the HIS classifying its own cells as *self* and everything else as *nonself* [5]. The HIS performs the classification by recognising proteins found on the surface of foreign cells (known as antigens). Foreign cells are different to cells present in the host (known as self-antigens) in structure and shape.

There are numerous instances however where this classification fails. For example, the intestinal tract is exposed to many different bacteria and food, neither of which are classically defined as `self', but neither of which produce an immune response. In addition, the model of self-nonself discrimination cannot explain the phenomena of auto-immune diseases. In the example of multiple sclerosis, the HIS attacks certain cells that it classifies as `self'. In 1994, Polly Matzinger [4] postulated that in this instance, the HIS was not reacting to self or nonself but was due to a protection mechanism of *sensing danger*. The manner in which danger is detected forms the basis of the Danger Theory.

The Danger Theory does not deny the existence of self-nonself discrimination but rather states there are other contributory factors involved in the initiation of an immune response. It is now believed that the HIS responds to certain danger signals produced as a result of cellular *necrosis*; the unexpected stress and/or death of a cell.

Cell death is a natural process that occurs within the body as a result of homeostatic regulation. This process however comes from a pre-programmed and highly controlled mechanism, known as *apoptosis*. The Danger Theory proposes that the mechanisms behind cell death can cause different biochemical reactions that in turn can cause different danger signals. It is believed that these signals may facilitate an immune response. This controversial paradigm shift within the immunology community may offer a potential explanation for many scenarios where the self-nonself model fails.

## ARTIFICIAL IMMUNE SYSTEMS

Most biologically inspired artificial immune systems based on the HIS have relied on the self-nonself model. Algorithms derived using this model have been largely successful [2]. Artificial immune systems have been developed for a wide range of applications from data mining to information security. In many cases, the applications have produced results comparable to, or better than, other standard techniques.

For example, the negative selection of immune cells in the thymus for self-nonself recognition was applied in the Lisys system and used as a network intrusion detection tool [3]. This system classified normal user behaviour as self and all other behaviour as nonself. However, this approach did not scale as well as expected for use in a large, dynamic environment. One explanation for the poor behaviour may be that certain processes, essential for immune functionality, were not incorporated.

## THE APPLICATION OF DANGER THEORY TO INTRUSION DETECTION

Intrusion detection systems (IDS) are designed to detect events that occur in a computer system that may compromise its integrity or confidentiality [7]. IDS are frequently sub-divided into two categories: *misuse detection* and *anomaly detection*. Misuse detection techniques examine both network and system activity for known instances of misuse through the use of signature matching algorithms. This technique is effective at detecting attacks that are already known. However, novel attacks are often missed giving rise to *false negatives*.

Anomaly detection systems rely on constructing a model of user behaviour that is considered 'normal'. This is achieved by using a combination of statistical or machine learning methods to examine network traffic or system calls and processes. The detection of novel attacks is more successful using the anomaly detection approach as any behaviour not defined as normal is classified as an intrusion. However, 'normal' behaviour in a large, dynamic system is not well defined and changes over time. This often results in a significant number of false alarms known as *false positives*. The reduction of false positives is a key challenge that the Danger Theory may be able to address.

It is proposed that the incorporation of the Danger Theory into intrusion detection techniques would produce a system able to respond effectively to known threats and novel attacks, and also reduce the amount of false positives common in anomaly detection systems. [6]. The Danger Theory proposes that the HIS detects danger signals and responds based on the correlation of these signals. A similar concept could be used in IDS. It would rely on being able to produce a system capable of classifying behaviour as *apoptotic* or *necrotic*. Apoptotic behaviour could be defined as low level, noisy alerts, which on their own do not form any significant misbehaviour, but are often the prerequisite for an attack. Necrotic alerts could be produced for a more serious attack where significant system damage was taking place [1]. Other danger signals relating to the physical system itself may also be incorporated into this model. The potential for improvement in this area and the successful correlation of such alerts will perhaps provide both improved intrusion detection systems and artificial immune systems.

## CONCLUSION

In the field of developing artificial immune systems for computer security, Danger Theory may provide significant improvements to current intrusion detection techniques. Work is currently being performed into exactly how danger signals can be identified in the HIS. It is hoped the results of this research will yield a clearer view on what danger signals are *in vivo*, how they can be translated for detecting danger within computer systems *in silico*, to implement more effective computer security systems.

## ACKNOWLEDGEMENTS

This project is a collaboration between the University of Nottingham, University Of The West Of England, University College London, Hewlett-Packard Labs, Bristol and Gianni Tedesco. The project is supported by the EPSRC (GR/S47809/01). Thanks to Gillan Cash for his helpful comments on this article.